\documentclass[10pt,letterpaper]{article}
\usepackage[top=0.85in,left=2.75in,footskip=0.75in,marginparwidth=1in]{geometry}

\usepackage[utf8]{inputenc}

\usepackage{cite}

\usepackage{nameref,hyperref}

\usepackage[right]{lineno}

\usepackage{microtype}
\DisableLigatures[f]{encoding = *, family = * }

\raggedright
\usepackage{changepage}

\usepackage[aboveskip=1pt,labelfont=bf,labelsep=period,singlelinecheck=off]{caption}

\makeatletter
\renewcommand{\@biblabel}[1]{\quad#1.}
\makeatother

\usepackage{lastpage,fancyhdr,graphicx}
\usepackage{epstopdf}
\fancyheadoffset[L]{2.25in}
\fancyfootoffset[L]{2.25in}

\usepackage{color}

\definecolor{Gray}{gray}{.25}

\usepackage{graphicx}

\usepackage{sidecap}

\usepackage{wrapfig}
\usepackage[pscoord]{eso-pic}
\usepackage[fulladjust]{marginnote}
\reversemarginpar

\begin{document}
\vspace*{0.35in}

\begin{adjustwidth}{-2.0in}{0in}

\begin{flushleft}
{\Large
\textbf\newline{Fusion of EEG and Musical Features in Continuous Music-emotion Recognition}
}
\newline
\\
Nattapong Thammasan\textsuperscript{1,*},
Ken-ichi Fukui\textsuperscript{2},
and Masayuki Numao\textsuperscript{2}
\\
\bigskip
\bf{1} Graduate school of Information Science and Technology, Osaka University, Osaka 565-0871, Japan
\\
\bf{2} Institute of Scientific and Industrial Research, Osaka University, Osaka 567-0047, Japan
\\
\bigskip
* nattapong@ai.sanken.osaka-u.ac.jp

\end{flushleft}

\begin{abstract}
	Emotion estimation in music listening is confronting challenges to capture the emotion variation of listeners. Recent years have witnessed attempts to exploit multimodality fusing information from musical contents and physiological signals captured from listeners to improve the performance of emotion recognition. In this paper, we present a study of fusion of signals of electroencephalogram (EEG), a tool to capture brainwaves at a high-temporal resolution, and musical features at decision level in recognizing the time-varying binary classes of arousal and valence. Our empirical results showed that the fusion could outperform the performance of emotion recognition using only EEG modality that was suffered from inter-subject variability, and this suggested the promise of multimodal fusion in improving the accuracy of music-emotion recognition.
\end{abstract}

\section{Introduction}\label{sec:introduction}

Recognizing human emotion during music listening is attracting widespread interest in the field of music information retrieval for many years~\protect\cite{yang_chen_2012} because it could enable a variety of application including music therapy, automatic music composition, and multimedia tagging. Since the early stage of this research area, musical features have been adopted due to the outstanding capability to reflect the \emph{expressed} emotion in music. Since the discovery of the relation between music-induced emotion and physiological patterns~\protect\cite{krumhansl_1997}, bodily signals directly recorded from listeners have been employed to model emotional response to music~\protect\cite{kim_andre_2008}. Among these attempts, an electroencephalogram (EEG), a tool to capture brainwaves, is a popularly adopted tool because of its excellent temporal resolution, cost effectiveness and fruitfulness of electrical activities nearby the brain, which is the center of emotion processing~\protect\cite{kim_review_2013}. 

In recent years, researchers have emphasized the importance of continuous emotion recognition over the course of time in response to multimedia stimuli~\protect\cite{Gunes_2013} (not limited to music stimuli). Automatic systems are expected to be responsive to user's time-varying emotion almost immediately. Recent works have been proposed to track time-varying emotion continuously annotated by users in response to music videos ~\protect\cite{Soleymani_2016} and songs~\protect\cite{thammasan_2016} using EEG dynamics. However, the performance was still limited owing to various challenges such as non-stationary of brain signals and disparity in EEG settings for different subjects.

Recent efforts to reinforce the emotion recognition model include using EEG features in conjunction with other information sources~\protect\cite{dmello_review_2015}, such as facial expression~\protect\cite{koelstra_fusion_2013}, and peripheral signals~\protect\cite{deap_2012,verma_fusion_2014}. One possible solution is to exploit information regarding the \emph{felt} emotion in conjunction with the \emph{expressed} emotion in music to estimate emotional state. In particular, a fusion of dynamic information from physiological signals and musical contents could possibly improve the performance of continuously estimating emotional response in music listening because both modalities could play a complementary role in music-emotion recognition model. Based on this concept, the only literature work (to our best knowledge) using EEG signals reported that the fusion of EEG dynamics and musical contents at feature level could improve music-emotion classification results~\protect\cite{Lin_Yang_fusion_2014}. Unfortunately, this work did not sufficiently take into account the time-varying characteristics of emotion during music listening as the methodology relied on emotion annotation with the granularity at musical-piece level. Therefore, the feasibility of using the fusion of EEG and musical features to improve continuous music-emotion recognition that considers emotion oscillation in music listening has not been proven.

In this paper, we present a study of multimodal fusion of EEG and musical features in the continuous emotion recognition. Features from each modality were fused at decision level (or late integration). Results of both subject-dependent and subject-independent emotion classification are presented. Furthermore, we also analyzed the effect of segmentation size, systematically investigated the contribution of each modality in this study.

To represent emotional state systematically, we adopted arousal-valence emotion model ~\protect\cite{russell_1980} that is one of the most commonly used models in the affective computing discipline. The model represents emotion in two continuous dimensions; arousal describes emotional intensity ranging from calm to activated emotion and valence describes positivity of emotion ranging from unpleasant to pleasant.

%
\section{Research Methodology}

\subsection{Experimental Protocol}

Twelve healthy male volunteers (averaged age = 25.59 y, SD = 1.69 y) were recruited to participate in our experiment. Each subject was instructed to select 16 songs from a 40-song music collection which is a set of MIDI files comprised of 40 instrumental pop songs having different instrument and tempo. The diversity of \emph{expressed} emotion and the balance of song familiarity in the selected songs were verified by the experimenter. Then, the songs were presented as synthesized sounds using the Java Sound API's MIDI package to the subject. By using MIDI files, any additional emotions contributed by lyrics can be eliminated. MIDI files also enable musical feature investigation and potential developing of music composition system which is considered as our future work. Songs in the library were between 73 to 147 s long (averaged length = 106.3 s, SD = 16.2 s). A 16 s silent resting period was inserted at the interval of each song to reduce any effect influenced by the previous song.

Simultaneously, EEG signals were acquired from the 12 electrodes of Waveguard EEG cap placed in accordance with the 10-20 international system. The positions of the selected electrodes were nearby the frontal lobe, which is believed to play a crucial role in emotion regulation~\protect\cite{Koelsch_nature_2014}. Throughout EEG recording, Cz electrode was used as a reference electrode and the impedance of each electrode was kept below 20 k$\Omega$. EEG signals were recorded at a 250 Hz sampling rate, amplified by Polymate AP1532 amplifier and visualized on APMonitor. A 0.5-60 Hz bandpass filter was also applied. A subject was also asked to keep his eyes close and minimize body movement during EEG recording to reduce any effect of unrelated artifacts. We also employed EEGLAB toolbox~\protect\cite{eeglab_2011} to remove eye-movement artifacts from the acquired EEG signals based on the independent component analysis (ICA) approach.

After music listening, EEG cap was removed from subject's scalp and the experiment proceeded to the emotion annotation session. In this session, a subject was instructed to annotate his \emph{felt} emotions in the previous session via our software. While listening to the same songs presented again in the same order, a subject reported the emotions by continuously clicking at a corresponding point in the arousal-valence emotion space shown on a monitor screen using a mouse. Arousal and valence were recorded independently as numerical values that ranged from --1 to 1. After providing an emotion annotation for each song, each subject was asked to confirm or change his familiarity with the song and indicate how confident, on a discrete scale ranging from 1 to 3, he was of the correspondence between the annotated emotions and the emotions perceived during the first listening phase.

\subsection{EEG Features}

In this work, we applied the fractal dimension (FD) approach to extract features from EEG signals due to its simplicity and excellent performance in previous affective computing studies~\protect\cite{sourina_real-time_2012,thammasan_2016}. Fractal dimension is a non-negative real value that quantifies the complexity and irregularity of data and can be used to reveal the complexity of a time-varying EEG signal. We applied Higuchi algorithm~\protect\cite{higuchi_approach_1988} to derive FD value from each particular window of EEG signals in this study.

Previous studies reported that asymmetries of features extracted from symmetric electrode pairs could be used as additional informative features to classify emotional states~\protect\cite{sourina_real-time_2012,thammasan_2016}. Therefore, we also added asymmetry indexes to our original EEG feature set by calculating the differential asymmetries of five left-right electrode pairs. All EEG features are summarized in Table~\ref{table_summary_feature}.

\subsection{Musical Features}

To extract emotion expression in music, we used the MIRtoolbox version 1.6.1~\protect\cite{mirtoolbox_2007}, which is a MATLAB toolbox that offers an integrated set of functions to extract musical features from audio files. Firstly, our MIDI files were converted into WAV format at a sampling rate of 44.1 kHz to be compatible to the toolbox. At a particular window, we subsequently extracted the high-level musical features using the \emph{mirfeatures} function.

A \emph{dynamic} feature of a song was derived from the frame-based root mean square of the amplitude (RMS) from the song. \emph{Rhythm} is the pattern of pulses/note of varying strength. We extracted the frame-based tempo estimation and the attack times and slopes of the onsets from songs. \emph{Timbre} reflects the spectro-temporal characteristics of sound. We extracted the spectral roughness that measures the noisiness of the spectrum, 13 Mel-frequency cepstral coefficients (MFCC) and their derivatives up to the $1^{st}$ order. In addition, we extracted the frame-decomposed zero-crossing rate, the low energy rate and the frame-decomposed spectral flux from songs. To extract \emph{tonal} characteristics, we calculated the frame-decomposed key clarity, mode, and the harmonic change detection function (HCDF) from songs.

Afterward, we calculated the means of the features of each window using the \emph{mirmean} function to overall represent the characteristic of the features in the window. The summary of musical features can be found in Table~\ref{table_summary_feature}. The features were selected by partly following the previous work~\protect\cite{Lin_Yang_fusion_2014}.

\subsection{Feature-level Multimodal Fusion of EEG and Musical Features}

\begin{table}[tbp]
	\begin{adjustwidth}{-2.0in}{0in}
	\centering
	\protect\caption{A summary of the extracted features}
	\label{table_summary_feature}
	
		\renewcommand{\arraystretch}{1.1}
		\scalebox{0.9}{
		\begin{tabular}{lcl}
			\hline \hline
			\textbf{Modality} & \multicolumn{1}{l}{\textbf{\# Features}} & \textbf{Extracted features}                                                                                                      \\ \hline
			EEG FD            & 12                                        & Fp1, Fp2, F3, F4, C3, C4, F7, F8, T3, T4, Fz, Pz                                  \\ \hline
			EEG FD Asymmetry & 5                                         & Fp1-Fp2, F3-F4, C3-C4, F7-F8, T3-T4                                                  \\ \hline
			Music Dynamic     & 1                                         & RMS                                                                                                                              \\ \hline
			Music Rhythm      & 3                                         & Tempo, Attack\_time, Attack\_slope                                                   \\ \hline
			Music Timbre      & 30                                        & \begin{tabular}[c]{@{}l@{}}Roughness, MFCC (1-13), dMFCC (1-13),\\ Zero-cross, Low\_energy, Spectral\_flux\end{tabular} \\ \hline
			Music Tonal       & 3                                         & Key\_clarity, Mode, HCDF                                                             \\ \hline \hline
		\end{tabular}
	}
	\end{adjustwidth}
\end{table}
%

In decision-level fusion, classification of each modality is processed independently and the output of classifiers are later combined to yield final results. In this work, we first classified EEG and music modalities individually and then combined the classifier outputs in a linear fashion.

For binary classification, let $p^x_{EEG}$ and $p^x_{music} \in[0,1]$ denote the classifier outputs of EEG and music modality respectively for class $x \in\{1,2\}$. Then the output class probability, namely $p^x_{multimodal}$, for class $x$ is given by
\begin{eqnarray}
p^x_{multimodal} = \alpha p^x_{EEG } + (1-\alpha)p^x_{music},
\label{eq:1}
\end{eqnarray}
\noindent where $\alpha$ is the weighting factor that satisfies $0\leq\alpha\leq1$ and determines how EEG modality contributes to the final decision. 

Although decision-level fusion allows asynchronous integration of different modalities, we used synchronous fashion by using the same window size for both EEG and music modality in order to allow a direct comparison between decision-level fusion and feature-level fusion. Similarly, we varied the size of sliding window from 2 to 10 s at a step of 1 s to investigate the effect of window size.

\subsection{Emotion Classification and Evaluation}

Despite the spatial continuity of arousal-valence space, most of recent attempts to estimate emotional states from EEG signals simply performed emotion recognition as classification rather than regression~\protect\cite{deap_2012,Lin_Yang_fusion_2014}. For the sake of simplicity, our work also addressed the binary emotion classification problem by categorizing valence into positive and negative classes and arousal into high and low arousal classes. Because of its success in literature~\protect\cite{kim_review_2013,Pachet_svm_rbf}, support vector machine (SVM) based on Gaussian radial basis kernel function (kernel scale = 3) was used to classify emotional classes. The SVM classifier was built by MATLAB Statistics and Machine Learning Toolbox\footnote{http://www.mathworks.com/products/statistics}.

Emotion classification model can be constructed in either subject-specific or generalized manner. In other words, the classification can be performed either dependently or independently to subjects. In this work, we investigated both strategies. In subject-dependent classification, stratified 10-fold cross-validation method was adopted to each subject's dataset, and the results of each individual were then averaged across subjects to derive overall performance. In subject-independent classification, we adopted leave-one-subject-out validation method to derive the performance of classification. In each trial, SVM classifier was trained with combined dataset from 11 subjects and then tested against the dataset from the remaining subject. Overall performance was computed by averaging across trials. Prior to classification, each feature was independently normalized to the range of $[0,1]$ using the min-max algorithm; we performed the normalization within a subject for subject-dependent classification and across all subjects for subject-independent classification.

Regarding a performance measurement, emotion classification accuracy was defined as the percentage of the correctly classified test instances in the total number of test instances. As self-reporting emotion annotation could lead to the imbalance in emotional classes. The unbalanced classes could mislead the implication of classification results, we, therefore, defined the \emph{chance level} as a new baseline. The \emph{chance level} of each subject was defined as the percentage of the number of instances in majority class in total instances. Both subject-dependent and subject-independent emotion classification results were compared to the chance levels to evaluate the relative performance of emotion recognition over majority-voting classification.

In addition to accuracy, we also used Matthews correlation coefficient (MCC)~\protect\protect\cite{mcc_1975}, which is a measure to reflect classification performance with consideration of class imbalance. MCC is a balanced measure and proper to be used even if the classes are of very different sizes. It reflects a correlation coefficient between the actual and the classified binary classes. The maximal coefficient +1 represents a perfect classification (100\% accuracy) and the minimal coefficient -1 represents total disagreement (0\% accuracy). The coefficient 0 indicates that the classification is one-class random guessing. Given a confusion matrix of binary classification, MCC can be calculated by
%
\begin{eqnarray}
\textstyle  
MCC = \frac{TP\times TN - FP\times FN}{\sqrt{(TP + FP)(TP + FN)(TN + FP)(TN + FN)}},
\end{eqnarray}
\noindent where $TP$ is the number of true positives, $TN$ is the number of true negatives, $FP$ is the number of false positives and $FN$ is the number of false negatives.
%
\section{Results}

We first investigated the results of subject-dependent and subject-independent classification by comparing decision-level fusion (DLF), EEG unimodality (EEG), music unimodality (MF) and chance level (Chance). In decision-level fusion, we used two different weighting factors ($\alpha$), 0.45 (DLF\_MF) and 0.55 (DLF\_EEG), to examine the effect of the weight difference on classification performance. Then, we further analyzed on decision-level fusion primarily focusing on the weighting factors. As some processes relied on randomization (10-fold cross-validation and the final decision of decision-level fusion), the classification was performed repeatedly for five times and we derived the average across all repetitions.

The averaged confidence level of correspondence in annotation across these remaining subjects was 2.4063 ($SD = 0.6565$), which indicated that the annotated data in our dataset was applicable. As familiarity was the main criteria in the song selection step, we found that song selection was diverse owing to different cultural backgrounds and musical preferences of subjects. The songs that were commonly selected by the majority of subjects was scarcely found.

\subsection{Results of Subject-dependent and Subject-independent Classification}

\begin{table*}[tbp]
	\begin{adjustwidth}{-2.0in}{0in}
	\centering
	\protect\caption{Averaged subject-dependent emotion classification accuracies across subjects}
	\label{tab:result_10_fold}
		\renewcommand{\arraystretch}{1.1}
		\setlength{\tabcolsep}{0.5pt}
		\scalebox{0.8}{
		\begin{tabular}{clccccccccc}
			\hline
			\hline
			Classification & Modality & \multicolumn{9}{c}{Window size (sec)}                                                                                                                                                                                \\ \cline{3-11} 
			&                           & 2                     & 3                     & 4                     & 5                     & 6                     & 7                     & 8                     & 9                     & 10                    \\ \hline
			
			Arousal        & DLF\_EEG & 82.88 (4.87)          & 82.81 (4.74)          & 82.64 (4.96)          & 81.9 (6.1)            & 81.75 (5.64)          & 81.19 (4.65)          & 80.33 (6.28)          & 80.43 (6.49)          & 79.94 (5.74)          \\ 
			& DLF\_MF  & 83.19 (4.4)           & 83.06 (4.69)          & 82.8 (5.45)           & 81.84 (5.38)          & 81.66 (5.39)          & 81.43 (5.08)          & 80.01 (6.23)          & 79.98 (6.09)          & 80.75 (5.58)          \\ 
			& EEG         & 82.12 (7.16)          & 82.08 (6.54)          & 81.49 (7.62)          & 81.01 (8.31)          & 80.9 (7.96)           & 80.55 (7.57)          & 79.89 (8.61)          & 79.37 (8.95)          & 78.94 (9.41)          \\ 
			& MF          & \textbf{84.17 (2.8)}           & \textbf{84.31 (3.4)}           & \textbf{83.96 (4.2)}           & \textbf{82.39 (4.16)}          & \textbf{82.38 (4.79)}          & \textbf{81.95 (3.74)}          & \textbf{81.13 (4.39)}          & \textbf{80.64 (4.95)}          & \textbf{81.05 (4.08)}          \\ 
			& Chance      & 62.48 (6.21)          & 62.67 (6.26)          & 62.4 (6.19)           & 62.67 (6.24)          & 62.52 (6.23)          & 62.43 (6.64)          & 62.58 (6.33)          & 62.4 (5.98)           & 62.42 (6.32)          \\ 
			\hline               
			Valence        & DLF\_EEG & 87.93 (5.92)          & 87.37 (5.79)          & 87.3 (5.77)           & 87 (6.06)             & 86.67 (5.91)          & 86.01 (6.38)          & 86.32 (5.69)          & 84.89 (6.22)          & 83.64 (6.73)          \\ 
			& DLF\_MF  & 88.34 (5.52)          & 87.9 (5.64)           & 87.87 (5.41)          & 87.84 (5.55)          & 86.74 (5.48)          & 86.49 (6.15)          & 86.66 (5.59)          & 85.5 (5.97)           & 84.64 (7.01)          \\ 
			& EEG         & 85.83 (7.7)           & 85.71 (7.71)          & 85.41 (7.55)          & 84.85 (7.86)          & 84.54 (7.72)          & 83.97 (8.3)           & 84.24 (7.88)          & 82.92 (7.91)          & 82.17 (8.65)          \\ 
			& MF          & \textbf{90.25 (4.73)}          & \textbf{89.53 (4.75)}          & \textbf{89.65 (4.83)}          & \textbf{89.37 (5)}             & \textbf{88.41 (4.79)}          & \textbf{88.47 (4.9)}           & \textbf{88.36 (4.62)}          & \textbf{87.57 (5.59)}          & \textbf{86.71 (5.7)}           \\ 
			& Chance      & 72.96 (12.67)         & 72.96 (12.66)         & 72.93 (12.7)          & 72.99 (12.76)         & 72.93 (12.73)         & 72.76 (12.79)         & 72.95 (12.93)         & 72.55 (12.95)         & 73.2 (12.9)           \\ \hline
			\hline               
		\end{tabular}
	}
	\end{adjustwidth}		
\end{table*}

\begin{figure}[t]
	\begin{adjustwidth}{-2.0in}{0in}
	\centering
	\includegraphics[width=1.0\linewidth]{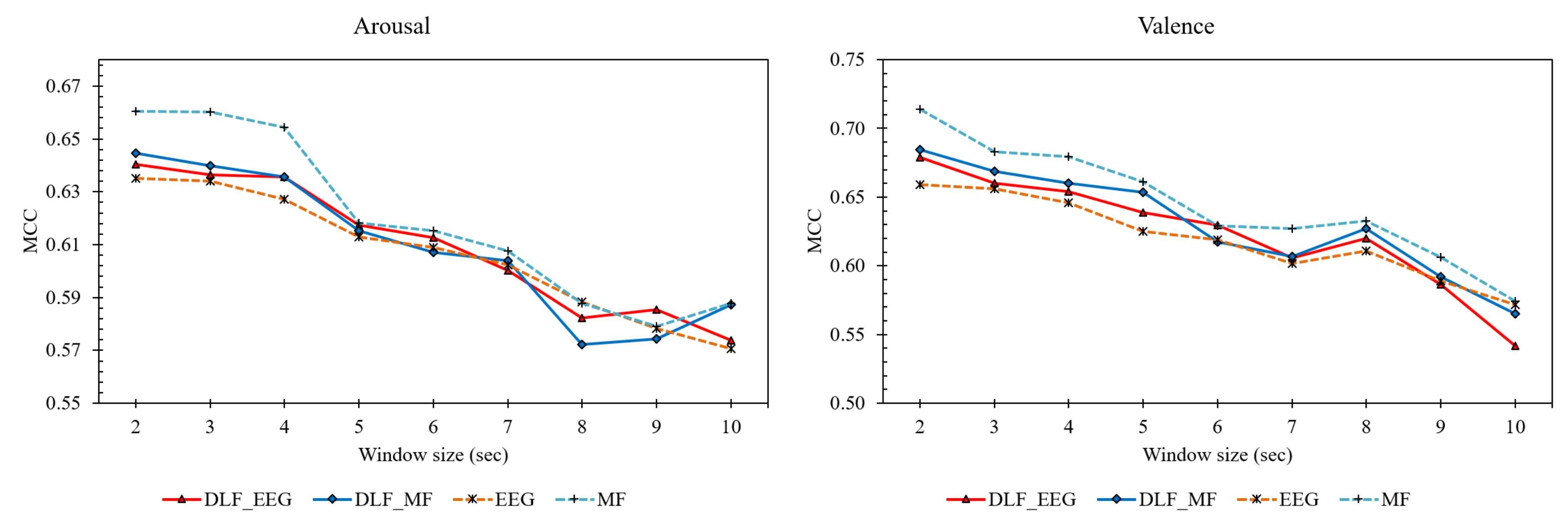}
	\protect\caption{Averaged subject-dependent emotion classification MCCs across subjects using different sliding window sizes}
	\label{fig:result_10_fold}
	\end{adjustwidth}
\end{figure}

\begin{table*}[tbp]
	\begin{adjustwidth}{-2.0in}{0in}
	\centering
	\protect\caption{Averaged subject-independent emotion classification accuracies across subjects}
	\label{tab:result_loso}
	
		\renewcommand{\arraystretch}{1.1}
		\setlength{\tabcolsep}{0.5pt}
		\scalebox{0.8}{
		\begin{tabular}{clccccccccc}
			\hline
			\hline
			Classification & Modality & \multicolumn{9}{c}{Window size (sec)}                                                                                                                                                                                \\ \cline{3-11} 
			&                           & 2                     & 3                     & 4                     & 5                     & 6                     & 7                     & 8                     & 9                     & 10                    \\ \hline
			
			Arousal        & DLF\_EEG & 55.93 (6.88)          & 56.66 (6.85)          & 55.41 (7.49)          & 56.33 (7.47)           & 56.64 (7.8)            & 56.89 (7.1)           & 56.86 (6.75)          & 56.89 (6.44)           & 56.2 (7.58)            \\ 
			& DLF\_MF  & 58.76 (6.73)          & 59.65 (6.68)          & 58.37 (6.71)          & 59.78 (7.24)           & 59.56 (7.62)           & 59.5 (7.06)           & 59.37 (6.49)          & 59.52 (6.16)           & 58.67 (7.71)           \\
			& EEG         & 44.09 (9.97)          & 44.05 (9.94)          & 43.7 (10.59)          & 43.99 (11.42)          & 45.17 (11.15)          & 44.68 (10.86)         & 44.6 (11.11)          & 44.72 (10.92)          & 44.88 (11.18)          \\ 
			& MF          & \textbf{71.08 (7.01)} & \textbf{72.18 (7.11)} & \textbf{70.42 (7.54)} & \textbf{72.34 (6.87)}  & \textbf{71.21 (7.43)}  & \textbf{71.82 (6.32)} & \textbf{70.86 (6.98)} & \textbf{71.54 (6.36)}  & \textbf{70.26 (8.03)}  \\ 
			& Chance      & 62.48 (6.21)          & 62.67 (6.26)          & 62.4 (6.19)           & 62.67 (6.24)           & 62.52 (6.23)           & 62.43 (6.64)          & 62.58 (6.33)          & 62.4 (5.98)            & 62.42 (6.32)           \\ 
			\hline
			Valence        & DLF\_EEG & 59.22 (10.13)         & 60.37 (10.02)         & 61 (10.33)            & 61.3 (10.35)           & 60.96 (9.75)           & 61.73 (10.27)         & 61.47 (9.48)          & 60.93 (10.81)          & 61.09 (9.95)           \\ 
			& DLF\_MF  & 60.97 (8.65)          & 61.79 (8.89)          & 61.95 (8.94)          & 63.21 (8.81)           & 62.78 (8.63)           & 63.02 (9.59)          & 62.86 (8.26)          & 63 (9.82)              & 63.51 (8.69)           \\ 
			& EEG         & 52.95 (15.77)         & 53.89 (16.23)         & 54.87 (16.63)         & 54.86 (16.65)          & 53.62 (16.16)          & 54.86 (16.03)         & 55.47 (15.87)         & 54.88 (16.74)          & 54.46 (16.3)           \\ 
			& MF          & \textbf{67.41 (6.6)}           & \textbf{68.7 (5.36)}           & \textbf{67.97 (6.36)}          & \textbf{70.1 (6.45)}            & \textbf{69.81 (5.12)}           & \textbf{70.39 (7.23)} & \textbf{69.24 (5.56)}          & \textbf{69.4 (5.51)}            & \textbf{70.4 (6.23)}            \\ 
			& Chance      & 72.96 (12.67)         & 72.96 (12.66)         & 72.93 (12.7)          & 72.99 (12.76)          & 72.93 (12.73)          & 72.76 (12.79)         & 72.95 (12.93)         & 72.55 (12.95)          & 73.2 (12.9)            \\
			\hline
			\hline               
		\end{tabular}
	}
	\end{adjustwidth}		
\end{table*}

\begin{figure}[t]
	\begin{adjustwidth}{-2.0in}{0in}
	\centering
	\includegraphics[width=1.0\linewidth]{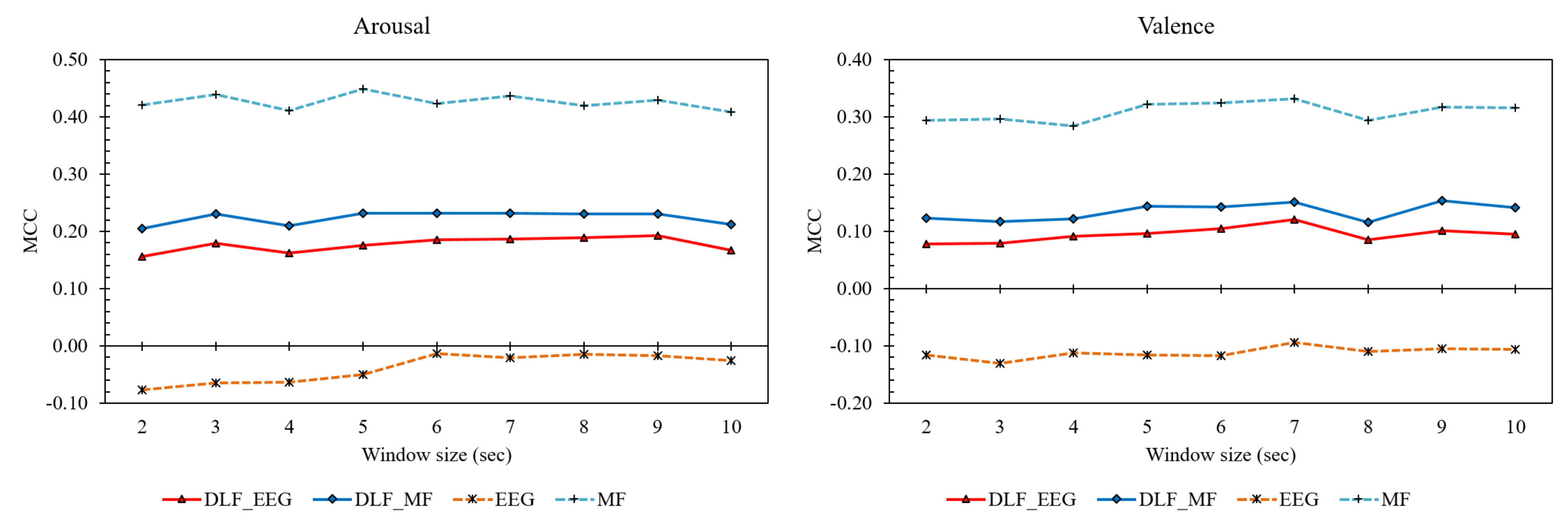}
	\protect\caption{Averaged subject-independent emotion classification MCCs across subjects using different sliding window sizes}
	\label{fig:result_loso}
	\end{adjustwidth}
\end{figure}

The averaged subject-dependent emotion classification accuracies across subjects using sliding windows with varied sizes are shown in Table~\ref{tab:result_10_fold} and the corresponding MCCs are illustrated in Figure~\ref{fig:result_10_fold}. According to the results, music unimodality achieved the best performance in both arousal and valence classification regardless of window size. Interestingly, fusing EEG modality with music modality outperformed other modalities in almost all of the cases. In general, decision-level fusion provided comparable results with unimodality. Interestingly, most of the modalities achieved their best performances when using sliding window size of 2 s.

Table~\ref{tab:result_loso} and Figure~\ref{fig:result_loso} summary the averaged subject-independent emotion classification accuracies and MCCs respectively. As can be seen, music modality achieved significantly better performance than other modalities. Interestingly, EEG modality provided the poorest results in every case. Our results suggested that the inter-individual variation in EEG signals may have a negative impact on emotion classification. Therefore, the inclusion of EEG signals could not improve the performance of subject-independent classification, and unimodality using musical features could be considered as more robust information to be employed in the construction of subject-independent emotion recognition model. Correspondingly, the decision-level fusion that relied slightly more on musical features than EEG features provided better results. In addition, the noticeable influence of sliding window size on classification performance could not be found.

\subsection{Analysis of Contribution of Each Modality in Decision-level Fusion}

\begin{figure*}
	\begin{adjustwidth}{-2.0in}{0in}
	\centering
	\includegraphics[width=1.0\linewidth]{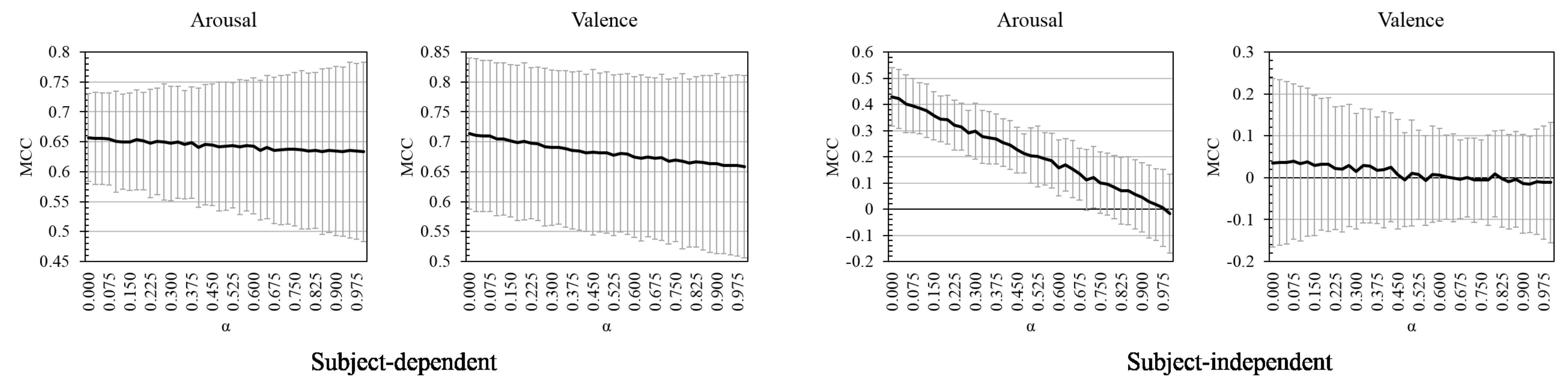}
	\protect\caption{Averaged emotion classification MCCs across subjects using decision-level fused features and fixed sliding window sizes with different weighting factors ($\alpha$ in Equation~\ref{eq:1}); the error bars represent the standard deviations}
	\label{fig:varying_alpha}
	\end{adjustwidth}
\end{figure*}

It was suggested from the literature~\protect\cite{deap_2012,koelstra_fusion_2013} and the above results that the difference in the contribution of each modality could influence results of decision-level fusion. We, therefore, further analyzed the effect of weighting factors ($\alpha$ in Equation~\ref{eq:1}) on classification in details by varying the factor from 0 (equivalent to music unimodality) to 1 (equivalent to EEG unimodality) at a step of 0.025. The sliding window size was fixed at 2 s for subject-dependent classification and 9 s for subject-independent classification because the sizes mainly achieved high performance in previous sections.

It can be observed from the results (Figure~\ref{fig:varying_alpha}) that the classification performance decreased when increasing the contribution of EEG features (namely varying $\alpha$ from 0 to 1), especially in subject-independent arousal classification. This suggested that music modality played more important role in emotion classification. Nevertheless, the higher variances at high $\alpha$ weighting factors in subject-dependent arousal classification indicated that EEG features could be more corresponding features to classify arousal classes in some subjects as well and thus provided better results.

\section{Discussion and Conclusion}

We have presented a study of multimodality using EEG and musical features in continuous emotion recognition. In this study we investigated on the varied sliding window size, subject-dependency of classification models, and the contribution of each modality. Empirically, EEG modality was suffered from the inter-subject variation of EEG signals and fusing music modality with EEG features could slightly boost emotion recognition. Future research is encouraged to study subjective factors in the variation and provide possible solution such as calibration or normalization over individuals. Nevertheless, the system cannot completely rely on the music unimodality based on the assumption that emotion in music listening is subjective. Completely discarding EEG modality would have adverse effects on practical emotion recognition model constructing.

Nevertheless, the results would infer to potential application in solving the \emph{cold start} problem. In particular, the emotion recognition system could use musical features to predict emotional states of a novel subject to the system at an initial state and then turn to use EEG features in conjunction with musical features to estimate emotion during music listening when the system is sufficiently reinforced by collecting more training data.

The acquired data has a limitation that leaves room for discussion. In particular, the class imbalance owing to self-annotation and the limited number of songs used for individual subject led us to apply merely the stratified 10-fold cross-validation despite the availability of leave-one-trial-out cross-validation. Future work should, therefore, focus on emotion scattering by either carefully controlling class balance in selected song or increasing the number of eliciting songs in order to enable another validation method. Apart from that, increasing the diversity of subjects, e.g. including female subjects, is also encouraged for future work.

In conclusion, we demonstrated that integrating musical features and EEG dynamics could be a promising approach to improve emotion classification.

%


\bibliographystyle{plain}
\bibliography{nips_reference}

\begin{thebibliography}{10}
	
	\bibitem{eeglab_2011}
	A.~Delorme, T.~Mullen, C.~Kothe, Z.A. Acar, N.~Bigdely-Shamlo, A.~Vankov, and
	S.~Makeig.
	\newblock {EEGLAB, SIFT, NFT, BCILAB, and ERICA}: New tools for advanced {EEG}
	processing.
	\newblock {\em Computational Intelligence and Neuroscience}, 2011, 2011.
	
	\bibitem{dmello_review_2015}
	S.K. D'mello and J.~Kory.
	\newblock A review and meta-analysis of multimodal affect detection systems.
	\newblock {\em ACM Computing Surveys}, 47(3):43:1--43:36, 2015.
	
	\bibitem{Gunes_2013}
	H.~Gunes and B.~Schuller.
	\newblock Categorical and dimensional affect analysis in continuous input:
	Current trends and future directions.
	\newblock {\em Image and Vision Computing}, 31(2):120--136, 2013.
	
	\bibitem{higuchi_approach_1988}
	T.~Higuchi.
	\newblock Approach to an irregular time series on the basis of the fractal
	theory.
	\newblock {\em Physica D}, 31(2):277--283, 1988.
	
	\bibitem{kim_andre_2008}
	J.~Kim and E.~Andre.
	\newblock Emotion recognition based on physiological changes in music
	listening.
	\newblock {\em IEEE Transactions on Pattern Analysis and Machine Intelligence},
	30(12):2067--2083, 2008.
	
	\bibitem{kim_review_2013}
	M.K. Kim, M.~Kim, E.~Oh, and S.P. Kim.
	\newblock A review on the computational methods for emotional state estimation
	from the human {EEG}.
	\newblock {\em Computational and Mathematical Methods in Medicine}, 2013, 2013.
	
	\bibitem{Koelsch_nature_2014}
	S.~Koelsch.
	\newblock Brain correlates of music-evoked emotions.
	\newblock {\em Nature Reviews Neuroscience}, 15(3):170--180, 2014.
	
	\bibitem{deap_2012}
	S.~Koelstra, C.~Muhl, M.~Soleymani, J.S. Lee, A.~Yazdani, T.~Ebrahimi, T.~Pun,
	A.~Nijholt, and I.~Patras.
	\newblock {DEAP}: A database for emotion analysis using physiological signals.
	\newblock {\em {IEEE} Transactions on Affective Computing}, 3(1):18--31, 2012.
	
	\bibitem{koelstra_fusion_2013}
	S.~Koelstra and I.~Patras.
	\newblock Fusion of facial expressions and {EEG} for implicit affective
	tagging.
	\newblock {\em Image and Vision Computing}, 31(2):164--174, 2013.
	
	\bibitem{krumhansl_1997}
	C.L. Krumhansl.
	\newblock An exploratory study of musical emotions and psychophysiology.
	\newblock {\em Canadian Journal of Experimental Psychology}, 51(4):336--353,
	1997.
	
	\bibitem{mirtoolbox_2007}
	O.~Lartillot and P.~Toiviainen.
	\newblock {MIR in Matlab (II)}: A matlab toolbox for music information
	retrieval.
	\newblock In {\em Proceedings of the 8th International Conference on Music
		Information Retrieval}, pages 127--130, 2007.
	
	\bibitem{Lin_Yang_fusion_2014}
	Y.P. Lin, Y.H. Yang, and T.P. Jung.
	\newblock Fusion of electroencephalogram dynamics and musical contents for
	estimating emotional responses in music listening.
	\newblock {\em Frontiers in Neuroscience}, 8(94), 2014.
	
	\bibitem{mcc_1975}
	B.W. Matthews.
	\newblock Comparison of the predicted and observed secondary structure of t4
	phage lysozyme.
	\newblock {\em Biochimica et Biophysica Acta (BBA) - Protein Structure},
	405(2):442--451, 1975.
	
	\bibitem{Pachet_svm_rbf}
	F.~Pachet and P.~Roy.
	\newblock Improving multilabel analysis of music titles: A large-scale
	validation of the correction approach.
	\newblock {\em IEEE Transactions on Audio, Speech, and Language Processing},
	17(2):335--343, 2009.
	
	\bibitem{russell_1980}
	J.A. Russell.
	\newblock A circumplex model of affect.
	\newblock {\em Journal of Personality and Social Psychology}, 39(6):1161--1178,
	1980.
	
	\bibitem{Soleymani_2016}
	M.~Soleymani, S.~Asghari-Esfeden, Y.~Fu, and M.~Pantic.
	\newblock Analysis of {EEG} signals and facial expressions for continuous
	emotion detection.
	\newblock {\em IEEE Transactions on Affective Computing}, 7(1):17--28, 2016.
	
	\bibitem{sourina_real-time_2012}
	O.~Sourina, Y.~Liu, and M.K. Nguyen.
	\newblock Real-time {EEG-based} emotion recognition for music therapy.
	\newblock {\em Journal on Multimodal User Interfaces}, 5(1--2):27--35, 2012.
	
	\bibitem{thammasan_2016}
	N.~Thammasan, K.~Moriyama, K.~Fukui, and M.~Numao.
	\newblock Continuous music-emotion recognition based on electroencephalogram.
	\newblock {\em {IEICE} Transactions on Information and Systems},
	E99-D(4):1234--1241, 2016.
	
	\bibitem{verma_fusion_2014}
	G.K. Verma and U.S. Tiwary.
	\newblock Multimodal fusion framework: A multiresolution approach for emotion
	classification and recognition from physiological signals.
	\newblock {\em NeuroImage}, 102, Part 1:162--172, 2014.
	
	\bibitem{yang_chen_2012}
	Y.H. Yang and H.H. Chen.
	\newblock Machine recognition of music emotion: A review.
	\newblock {\em ACM Transactions on Intelligent Systems and Technology},
	3(3):40:1--40:30, 2012.
	
\end{thebibliography}

	\end{adjustwidth}


\nolinenumbers



\end{document}